\documentclass[journal, letterpaper, final]{IEEEtran}

\usepackage{times}
\usepackage{epsfig}
\usepackage{graphicx}
\usepackage{amsmath}
\usepackage{amssymb}
\usepackage{multirow}
\usepackage{caption}
\usepackage{subcaption}
\usepackage{mathtools}
\usepackage{verbatim}
\usepackage{color}
\usepackage{float}
\usepackage[lined,boxed,commentsnumbered]{algorithm2e}
\usepackage{pbox}
\usepackage{afterpage}

\newcommand{\eqb}[1]{\begin{equation}\label{#1}}
\newcommand{\eqe}{\end{equation}}
\newcommand{\splitb}{\begin{align}\begin{split}}
\newcommand{\splite}{\end{split} \end{align}}

\newcommand{\matb}{\left( \begin{matrix*}[r] }
\newcommand{\mate}{\end{matrix*}\right)}

\DeclareMathOperator*{\argmax}{arg\,max}

\newcommand{\Section}[1]{\vspace{-4pt}\section{#1}\vspace{-4pt}}

\newcommand{\Subsection}[1]{\vspace{-3pt}\subsection{#1}\vspace{-3pt}}

\usepackage[breaklinks=true,bookmarks=false]{hyperref}

\graphicspath{{Figures/}}

\begin{document}

\title{Fast Sublinear Sparse Representation using Shallow Tree Matching Pursuit}

\author{Ali~Ayremlou, ~\IEEEmembership{Member, ~IEEE, } Thomas~Goldstein, ~\IEEEmembership{Member, ~IEEE} Ashok~Veeraraghavan, ~\IEEEmembership{Member, ~IEEE} and ~Richard~Baraniuk,~\IEEEmembership{Fellow,~IEEE}
\thanks{A. Ayremlou, T. Goldstein, A. Veeraraghavan and R. Baraniuk are with the Department of Electrical and Computer Engineering, Rice University, Houston, TX 77005 USA (e-mail: \{a.ayremlou, tag7, vashok, richb\}@rice.edu).}}

\maketitle

\begin{abstract}
Sparse approximations using highly over-complete dictionaries is a state-of-the-art tool for many imaging applications including denoising, super-resolution, compressive sensing, light-field analysis, and object recognition.
Unfortunately, the applicability of such methods is severely hampered by the computational burden of sparse approximation: these algorithms are linear or super-linear in both the data dimensionality and size of the dictionary.

We propose a framework for learning the hierarchical structure of over-complete dictionaries that enables fast computation of sparse representations.  Our method builds on tree-based strategies for nearest neighbor matching, and presents domain-specific enhancements that are highly efficient for the analysis of image patches.  Contrary to most popular methods for building spatial data structures, out methods rely on {\em shallow}, balanced trees with relatively few layers.
    
We show an extensive array of experiments on several applications such as image denoising/superresolution, compressive video/light-field sensing where we practically achieve 100-1000x speedup (with a less than 1dB loss in accuracy).

\end{abstract}

\Section{Introduction}
\label{sec:Intro}
\IEEEPARstart{C}{ompressive} sensing and sparse approximation using redundant
dictionaries are important  tools for a wide range of imaging applications including
image/video denoising \cite{elad2006image, protter2009image, mairal2007learning}, superresolution \cite{yang2010image, kong2006video}, compressive sensing of videos \cite{wakin2006compressive, reddy2011p2c2, hitomi2011video}, light-fields \cite{marwah2013compressive, salahieh2013compressive}, hyperspectral data \cite{li2013hyperspectral}, and even inference tasks such as face and object recognition \cite{ jiang2013label}. 
\begin{figure}[ht]
   \centering
      \includegraphics[scale = .45]{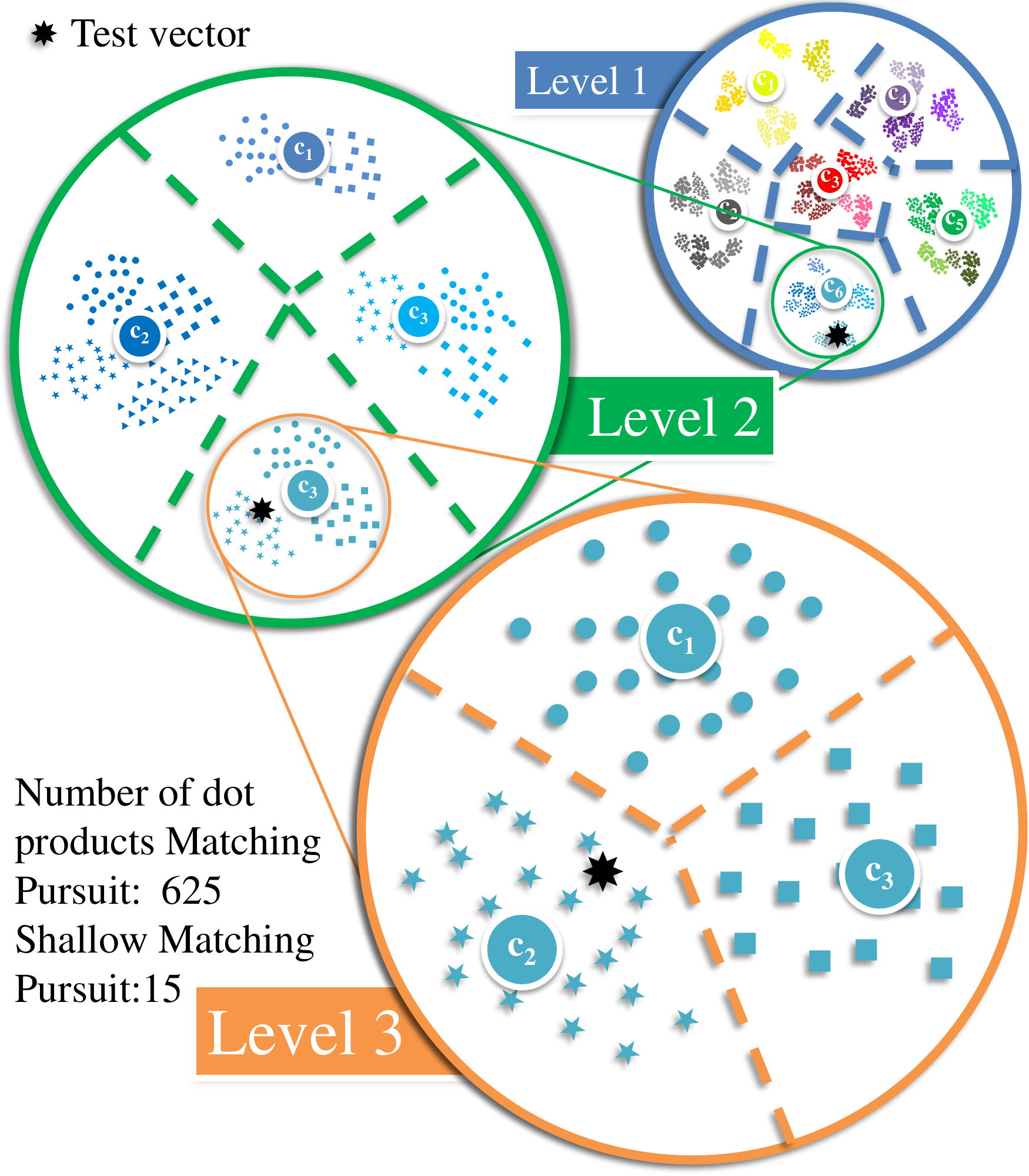}
   \caption{Shallow Tree representations match a test vector(black star) to a dictionary by traversing a shallow (3 level) tree in which each node may have many children.}
\label{fig:Overview}
\end{figure}

In spite of this widespread adoption in research, their adoption in commercial and practical systems is still lacking.
One of the principal reasons is the computational complexity of the algorithms needed: these algorithms are either linear or super-linear in both the data dimensionality and size of the dictionary. Common applications requiring dictionaries with over $10^5$  atoms require computation times that may exceed several days.

As an illustrative example, consider the problem of compressive video sensing using overcomplete dictionaries \cite{hitomi2011video}.
In \cite{hitomi2011video}, an overcomplete dictionary consisting of $10^5$ video patches was learned and utilized for compressive sensing using orthogonal matching pursuit (OMP).
Reconstruction for a single video (36 frames) using $10^5$ dictionary atoms takes more than a day, making these methods impractical for most applications and highlighting the need for significantly faster algorithms.

\Subsection{Motivation and Related Work}

Algorithms for sparse approximation can be broadly classified into two categories: those based on convex optimization, and greedy pursuit algorithms \cite{pati1993orthogonal}.
Several attempts have been made to develop fast algorithms for sparse regularization based on convex optimization 
\cite{figueiredo2007gradient, spgl1:2007, hale2008fixed}.
In spite of the progress made, these algorithms may still be slow when the size of the dictionary and data dimensionality become large.


\textbf{Fast Matching Pursuit:} Matching Pursuit(MP) and its many variants \cite{pati1993orthogonal}\cite{gribonval2001fast}  build sparse approximations by sequentially choosing atoms from a dictionary, and adding them one-at-a-time to the current ensemble.  On each stage, the target vector is approximated using the current ensemble, and the approximation error or ``residual'' is measured.  Next, an atom is selected from the dictionary that best approximates the current residual.
The computational bottleneck of this process is finding the dictionary atom closest to the residual.
Nearest Neighbor(NN) search methods face a similar bottleneck that has been aggressively tackled using Approximate Nearest Neighbor (ANN) search algorithms \cite{arya1998optimal,andoni2006near,muja2009fast}.
Most ANN search methods organize data into a tree structure that enables fast retrieval at query time  \cite{muja2009fast} . 
Typically a very deep tree with binary branching at every level is learned.

Hierarchical/tree approaches have been used in many applications to speed up dictionary matching. In \cite{bo2011hierarchical}, Batch Tree Orthogonal Matching Pursuit (BTOMP) is used to build a feature hierarchy that yields a better classification.  The authors of \cite{5995719} construct trees using  ``kernel descriptors'' for the same application.  Hierarchical methods for representing image patches are studied in \cite{5995732}, \cite{bo2012multipath}.The authors of \cite{6409355}  Random prejections for dimensionally reduction are used in \cite{xiang2011learning} to build hierarchical dictionaries. In \cite{4358654} binary hierarchical structure and PCA (Principal Component Analysis) are combined to reduce the complexity of the OMP.

Unfortunately, such a deep tree does not provide a beneficial trade-off between accuracy and speedup for dictionaries, since these atoms tend to be highly coherent.
Further, because they require backtracking and branch-and-bound methods, typical ANN techniques such as kd-trees do not provide reliable runtime guarantees.
 
In contrast, we organize the dictionary using a shallow tree (typically 3 levels) as shown in Figure \ref{fig:Overview}.
Our tree construction scheme is such that the resulting tree represents a balanced, hierarchical clustering of the atoms. 
Finally, we devise a sublinear time search algorithm for identifying the support set that provides the user with precise control over the computational speedup achieved while retaining high fidelity approximations.

\Subsection{Contributions}
 We propose an algorithm for balanced hierarchical clustering of dictionary atoms. We exploit the clustering to derive a sublinear time algorithm for sparse approximation. Our methods has a single parameter $\alpha$ that  provides fine-scale control on the computational speed-up achieved, enabling a natural trade-off between accuracy and computation.  We perform extensive experiments that span numerous applications where shallow trees achieve 150-1000x speedup (with a 1dB of less loss in accuracy) compared to conventional methods.


\Section{Problem Formulation}
\label{sec:Prob}
\Subsection{Sparse Approximation using Dictionaries}
Our approach to fast dictionary coding uses  Matching Pursuit (MP), which is a greedy method for selecting the constituent dictionary elements that comprise a test vector.   MP is a commonly used scheme for this application because dictionary representations of image patches are extremely sparse.  
For computing representations involving large numbers of atoms (e.g. for representing entire images rather than just patches) more complex pursuit algorithms have been proposed \cite{NT08} that we do not consider here.    

\textbf{Matching Pursuit:}
MP is a stage-wise scheme that builds a signal representation one atom at a time.   Algorithm \ref{MP_alg} is initialized by declaring the ``residual'' $r$ to be equal to the test vector $x$.  This residual represents the component of $x$ that has not yet been accounted for by the sparse approximation. In each iteration of the main loop an atom enters the representation.  The atom is selected by computing inner products with all (normalized) columns in $D = \{d_i\}$ and selecting the atom with the largest inner product.  The residual is then updated by subtracting the contribution of the entering dictionary element.

\begin{algorithm}
\DontPrintSemicolon
\KwData{Normalized Dictionary $D\in\mathbb{R}^{n\times m}$, Input vector $x\in \mathbb{R}^{n}$, Sparsity level $K\in \mathbb{Z}^{+}$}
\KwResult{Sparse vector $s\in \mathbb{R}^{m}$ with $x\approx Ds$}
\Begin{
      $r = x$\;
    \tcc{Main Loop}
    \For{$k = 1$ \KwTo $K$}{
        $ i \leftarrow \argmax_i   | d_i \cdot x | $ \;
        $s_i = d_i\cdot x$\;
        $r  \leftarrow r - d_i s_i$\;
    }
}
\caption{Maching Pursuit}\label{MP_alg}
\end{algorithm}

\textbf{MP Computational Complexity:}
MP requires the computation of $m$ inner products on the ``matching'' stage of each iteration.  Since each inner product requires $O(n)$ operations and there are $K$ stages, the overall complexity is $O(mnK).$   Note that this complexity is dominated by $m,$ the number of atoms in the dictionary.  For most imaging applications, $D$ is highly over-complete.  A typical image denoising method may operate on $16\times 16$ image patches ($n =16\times16=256$), use $K=5$ atoms per patch, and require $m=100,000$ dictionary elements.  For video or light field applications, $m$ may be substantially larger.  the computational burden of handling large dictionaries is a major roadblock for use in applications.    

\Subsection{Problem Definition and Goals}
   We consider variations on MP that avoid the brute-force $O(mn)$ matching of dictionary elements.  Our method is based on a hierarchical clustering method that organizes an arbitrary dictionary into a tree.  The tree can be traversed efficiently to approximately select the best atom to enter the representation on each stage of MP. 
   Our method is conceptually related to ANN methods (such as k-d trees).   However, unlike conventional ANN schemes, the proposed method is customized to the problem of dictionary matching pursuit, and so differs from conventional ANN methods in several ways.  The most significant difference is that the proposed method uses ``shallow'' trees (i.e. trees with a very small number of layers), as opposed to most ANN methods with use very deep trees with only two branches per level.

\Section{Algorithm for Hierarchical Clustering}
\label{sec:Cluster}
The proposed method relies on a hierarchically clustering that organizes dictionaries trees.  Each node of the tree represents a group of dictionary elements.  As we traverse down the tree, these groups are decomposed into smaller sub-groups.   To decompose groups of atoms into intelligent components, we use an algorithm based on k-means.  To facilitate fast searching of the resulting tree, we require that each node be balanced -- i.e., all nodes at the same level of the tree represent the same number of atoms.   Conventional k-means, when applied to image dictionaries, tends to produce highly unbalanced clusters, sometimes with as many as 90\% of atoms in a single cluster.  For the purpose of tree search, this is clearly undesirable as descending to this branch of the tree does not substantially reduce the number of atoms to choose from.  For this reason, the proposed clustering uses ``balanced'' k-means.

\Subsection{Balanced Clustering} \label{sec:balanced}
We now consider the problem of uniformly breaking a set of elements into smaller groups.   We begin with a collection of $m$ atoms to be decompose into $k$ groups.  We apply k-means to the atoms.   We then examine only the ``large'' clusters that contain at least $\lfloor m/k \rfloor$ atoms and discard the rest.  For each large cluster, we keep the  $\lfloor m/k \rfloor$ nearest atoms to the centroid, and discard the remaining atoms.   Suppose $\hat k$ such clusters are identified.  The algorithm is then repeated by applying k-means to the remaining unclustered atoms to form $k-\hat k$ groups. Once again, groups of at least  $\lfloor m/k \rfloor$ atoms are identified, and reduced to a cluster of exactly  $\lfloor m/k \rfloor$ elements.  This process is repeated until the number of remaining atoms is less than  $\lfloor m/k \rfloor$, at which point the remaining atoms form their own last cluster.

\Subsection{Hierarchical Clustering} \label{sec: cluster}
The clustering method in Section \ref{sec:balanced} can be used to organize dictionaries into hierarchical trees.   We begin with a parent node containing all dictionary atoms. The dictionary is then decomposed into $k_1$ balanced groups.  Each such groups is considered to be a ``child'' of the parent node.  Each child node is examined, and the atoms it contains are partitioned into $k_2$ groups which become its children.  This process is repeated until the desired level of granularity is attained in the bottom-level (leaf) nodes of the tree.

\Subsection{Fast Matching using Shallow Tree}
Using the tree representation of the dictionary, ANN matches can be found from a given test vector $x$.  The goal is to find the dictionary entry with the largest inner product with $x$.  The tradeoff between precision and speed is controlled by a parameter $\alpha$.  The search algorithm begins by considering the top-level node in the tree.  The test vector $x$ is compared to the centroid of the cluster that each child node represents.  Using the notation of Section \ref{sec: cluster}, there are $k_1$ such clusters.  We retain the $\lceil \alpha k_1 \rceil$ clusters with the largest inner products with the test vector.  The search process is then called recursively on these nearby clusters to approximately find the closest atom in each cluster.  The resulting  $\lceil \alpha k_1 \rceil$ atoms are compared to $x$, and the closest atom is returned.

\begin{algorithm}[h]
\DontPrintSemicolon
\textbf{Function} STMP($N$, $x$)   \;
\KwData{Tree Node $N$,  Input vector($x\in \mathbb{R}^{n}$) }
\KwResult{Approximate nearest dictionary atom}
\Begin{
    Let $\{C_i\}_{i=1}^k$ denote $N$'s children\;
   Retrieve centroids $\{c_i\}_{i=1}^k$ of $\{C_i\}$ \;
     \If{ $\{C_i\}$ have no children }{\textbf{Return} $c_i$ that maximizes $| x \cdot c_i | $ }
   Let $\{C_i^{min}\}_{i=1}^{\lceil \alpha k \rceil}$ denote the $\lceil \alpha k \rceil$ children with maximal $|c_i^{min} \cdot x|$ \;
   \textbf{Return} $c_i^{min}$ that maximizes $|x \cdot \text{STMP}(C_i^{min},x) |$
}
\caption{Shallow Tree Matching Pursuit}\label{BB_alg}
\label{alg:match}
\end{algorithm}

\begin{figure}
   \centering
      \includegraphics[width = \linewidth]{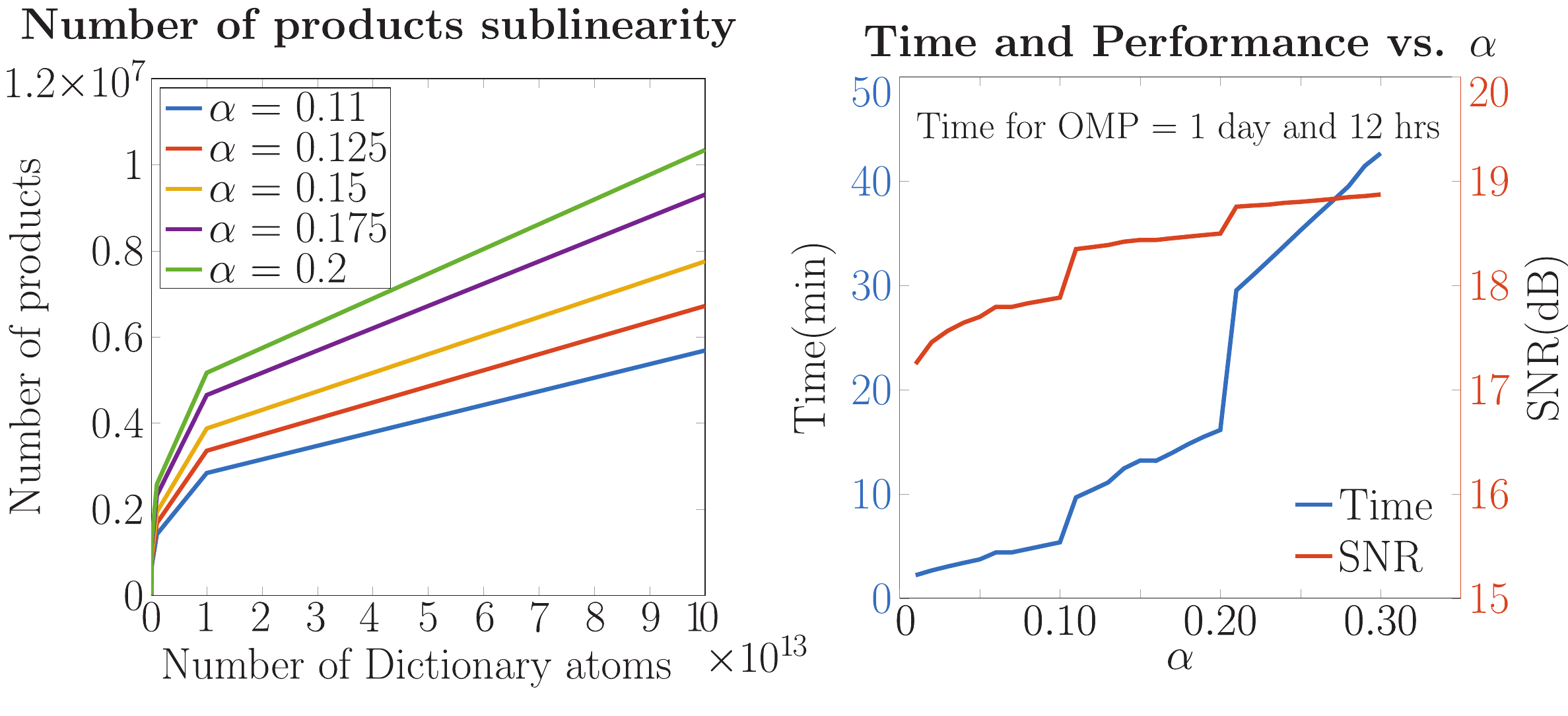}
   \caption{The number of inner products needed for matching scales sublinearly with dictionary size.}
\label{fig:complexity}
\end{figure}

\textbf{Computational Complexity:}
   At the first level of the tree, Algorithm \ref{alg:match} must compute $k_1$ inner products.  On the second level of the tree, $\alpha k_1 k_2$ inner products are computed, and $\alpha^2 k_1 k_2 k_3$ on the third, etc.  It total, the number of inner products is given by
    $\sum_{i=1}^L \alpha^{i-1}\prod_{j=1}^i k_j.$
  As long as $k$ remains bounded and $\alpha<1$, this grows sub-linearly with $m$.  In particular, if we choose $k_i = 1/\alpha$ for all $i>i^*,$ then the number of inner products is $O(\log(m))$ and the total complexity (including the cost of inner products) is $O(n\log(m)).$ Figure \ref{fig:complexity} shows the inner products needed to match an atom for a variety of parameter choices and dictionary sizes.  Note the sublinear scaling with dictionary size.

  \textbf{Construction of Shallow trees:}
  For all experiments in this article we use trees with only 3 levels. We choose $k_1=100,$  $k_2=k_3=10,$ and $\alpha=0.1.$  Because we have chosen $k_2,k_3=1/\alpha,$ the number of inner products that are computed does not grow as we descent lower into the tree. 
  
    We call the proposed method a ``shallow tree'' algorithm because the hierarchical clustering generates trees with only 3 levels and 100 branches on the first node.  This is in sharp contrast to conventional tree-based nearest neighbor methods (see e.g., \cite{andoni2006near}) that rely on very deep trees with only two branches per node.  For use with image dictionaries, shallow trees appear to perform much better for patch matching than conventional off-the-shelf nearest neighbor methods.



\section{Experimental Results}
\label{sec:ExpDraft}
We compare Shallow Tree Matching Pursuit with other algorithms in terms of both run-time and reconstruction quality for a variety of problems.
The main conclusion from the experiments is that STMP provides a 100-1000x speedup compared to existing sparse regularization methods with less than $1 dB$ loss in performance.

\begin{figure*}[!htb]
\centering
\includegraphics[scale = 0.6]{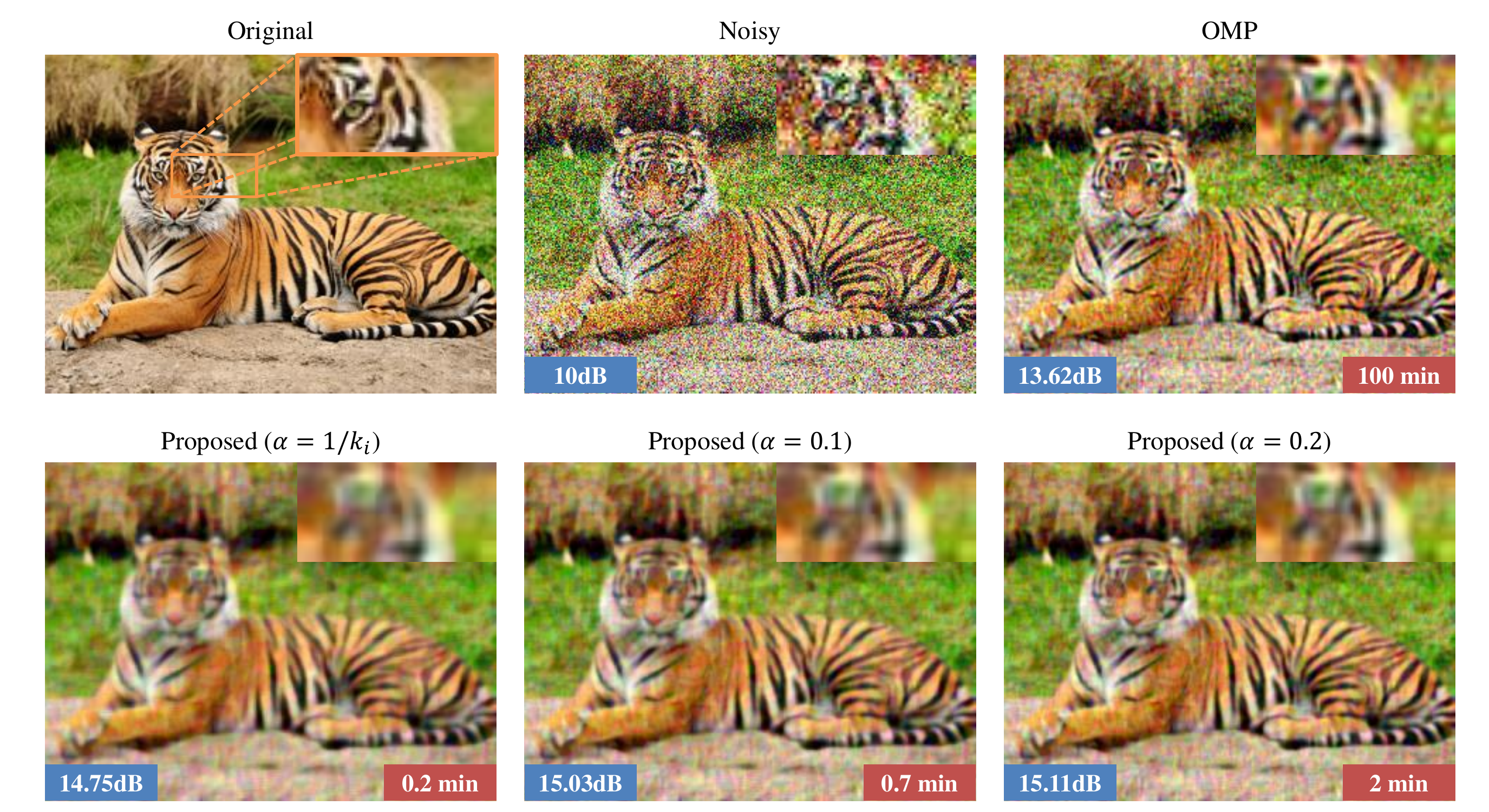}
\caption{Sample results for image denoising compared to other methods.}
\label{fig:ImgDe}

\includegraphics[scale = 0.3]{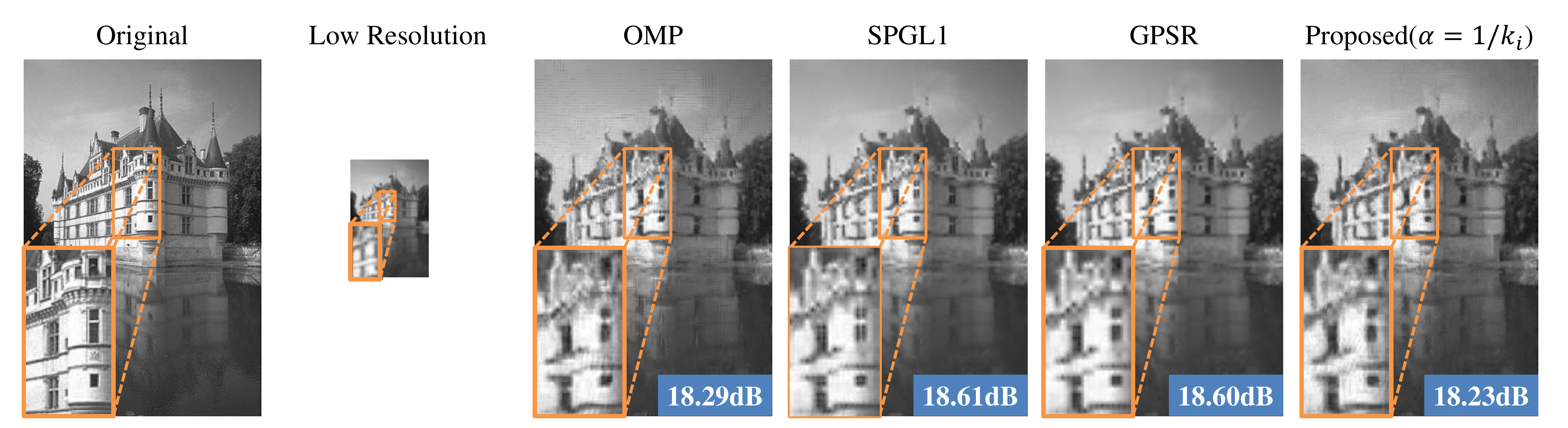}
\caption{Image Super-Resolution(4x): Proposed method has same performance with significantly faster run-times.}
\label{fig:ImgSupRes}
\end{figure*}

\begin{table*}[!htb]
\begin{center}
\begin{tabular}{| l |c|c|c|c|c|c|c|c|c|c|c|c|c|c|c|}
\cline{3-16}
\multicolumn{1}{ l  }{}& \multicolumn{1}{ c|  }{}&\multicolumn{7}{|c|}{\bf{Denoising}}  & \multicolumn{7}{|c|}{\bf{Super-Resolution(4x)}}\\

\cline{3-16}
\multicolumn{1}{ l }{}& \multicolumn{1}{ c|  }{}&\multicolumn{2}{|c|}{\bf{Berkeley1}}  & \multicolumn{2}{|c|}{\bf{Berkeley2}}  & \multicolumn{2}{|c|}{\bf{Berkeley3}}& & \multicolumn{2}{|c|}{\bf{Berkeley1}}  & \multicolumn{2}{|c|}{\bf{Berkeley2}}  & \multicolumn{2}{|c|}{\bf{Berkeley3}}&\\

\cline{1-8}\cline{10-15}
&Dic.          & PSNR      & Time                 & PSNR      & Time              & PSNR      & Time  & \bf{Run} & PSNR      & Time                & PSNR      & Time               & PSNR      & Time       &   \bf{Run}               \\

\bf{Method}  &Size          & (dB)      & (min)                 & (dB)      & (min)               & (dB)      & (min) & \bf{Time}  & (dB)      & (min)                 & (dB)      & (min)               & (dB)      & (min)       &   \bf{Time}      \\

\hline\hline
Proposed ($\alpha = 1/k_{i}$)   & 40k& 18.55         & 0.07                      & 20.46         & 0.07                    & 17.02         & 0.07  &1x 
& 17.16         & 0.02                      & 18.23         & 0.02                    & 14.96         & 0.02            & 1x             \\

Proposed ($\alpha = 0.1$)   &40k& 18.79         & 0.21                      & 20.70         & 0.21                    & 17.76         & 0.21 &3x
& 17.42         & 0.04                      & 18.51         & 0.04                    & 15.15         & 0.04            &  2x           \\


\hline
OMP        &40k             & 18.37         & 21.41                     & 19.89         & 21.46                   & 19.27         & 17.37  & 286x 
& 16.94         & 0.55                     & 18.29         & 0.56                   & 15.08         & 0.57             &  28x          \\


\hline
SPGL1 &4k               & 14.92         & 95.81                     & 16.06         & 90.42                   & 16.34         & 80.96    & 1272x
& 17.27         & 13.88                     & 18.61         & 13.00                   & 15.30         & 13.52             & 673x          \\

FPC\_AS& 4k             & 13.64         & 124.2                     & 14.85         & 123.0                   & 14.85         & 123.0       & 1763x
& 17.24         & 2.44                     & 18.61         & 2.58                   & 15.30         & 2.56        & 126x         \\

GPSR &4k                & 18.72         & 30.29                     & 20.40         & 26.55                   & 19.36         & 39.81    & 460x   
& 17.24         & 11.37                     & 18.60         & 9.56                   & 15.26         & 28.04        & 816x         \\

\hline
KD-Tree&40k                 & 18.06         & 101.7                    & 19.63         & 101.7                    & 17.21         & 101.7    &  1452x 
& 16.77         & 2.66                   & 17.96         & 2.22                    & 14.68         & 2.23         & 33x             \\

ANN     &40k                & 18.03         & 24.58                     & 19.59         & 25.41                   & 16.60         & 25.29    &358x   
& 16.77         & 0.83                     & 17.96         & 0.81                   & 14.68        & 0.83      & 11x     \\
\hline

\end{tabular}
\end{center}
\caption{Run-Time comparisons for Image Denoising and Image Super-Resolution(4x): STMP is 100-1000x faster with a $1$ dB loss in PSNR.}
\label{table:ImgDe}
\end{table*}

We compare to the following techniques:

{\textbf{STMP:} Shallow Tree Matching Pursuit with three different values of $\alpha$, i.e., $\alpha=\frac{1}{N},\alpha=0.1$ and $\alpha =0.2$. Lower $\alpha$ results in faster run-time, while larger $\alpha$ results in better approximations. Our implementation is in matlab.}

{\textbf{OMP:} Orthogonal Matching Pursuit (OMP) is a popular pursuit algorithm used in several vision applications. We use the mex implementation available as a part of the K-SVD software package \cite{rubinstein2008efficient}.}

{\textbf{SPGL1:} A matlab solver for large scale $L_1$ regularized least squares problems \cite{spgl1:2007}. This code achieves sparse coding via basis pursuit denoising problems.}

{\textbf{FPC-AS:} A matlab solver for $L_1$ regularized least squares based on fixed point continuation \cite{hale2008fixed}.  Due to impractically slow performance FPC-AS is only tested in imaging problems.}

{\textbf{GPSR:} A matlab solver for sparse reconstruction using gradient projections \cite{figueiredo2007gradient}.}

{\textbf{KD-Tree:} We use the built-in matlab function for fast approximate nearest neighbor search using kd-trees to speed up traditional matching pursuit.}

{\textbf{ANN:} We use the ANN C++ library for approximate nearest neighbor matching to speed up traditional matching pursuit \cite{arya1998optimal}.}

\textbf{Other Notes:} We sometimes use a smaller randomly sub-sampled dictionary to test the variational methods in cases where runtimes were impractically long ($>24$ hours).
All experiments begin by breaking datasets into patches using a sliding frame.  A restored image is then synthesized by averaging together the individual restored patches.


\Subsection{Imaging Experiments}

\begin{figure*}[!htb]
\centering
\includegraphics[scale = 0.6]{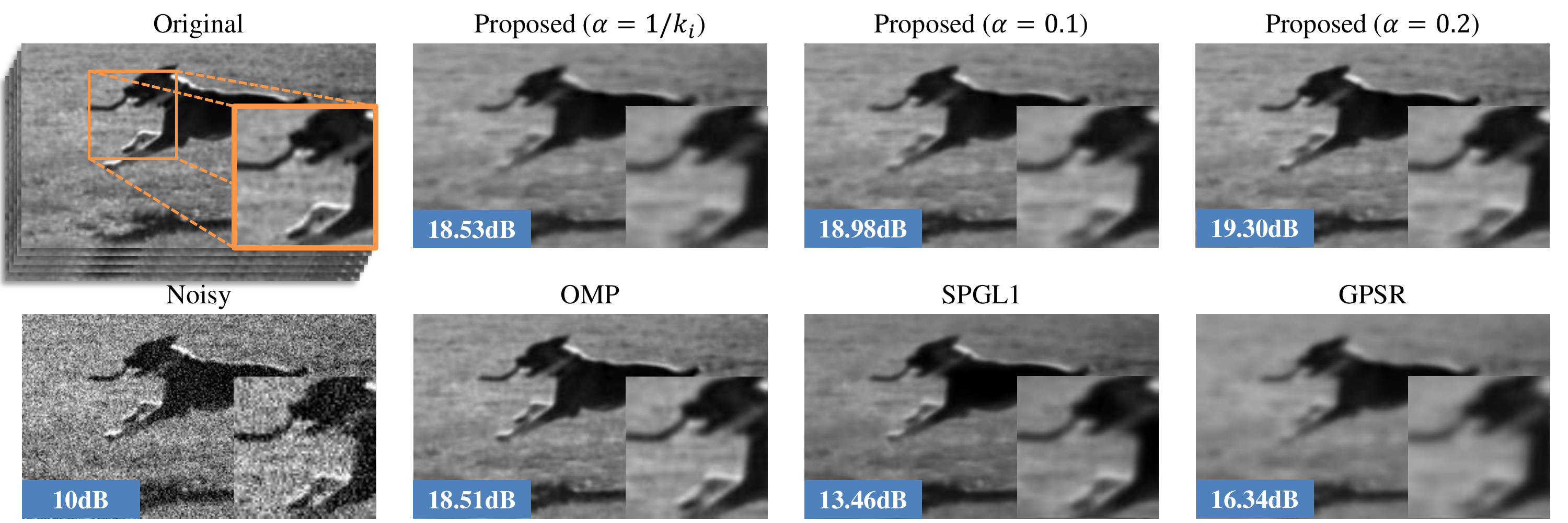}
\caption{Sample frame shown for video denoising application in comparison to other methods}
\label{fig:VidDe}
\includegraphics[scale = 0.35]{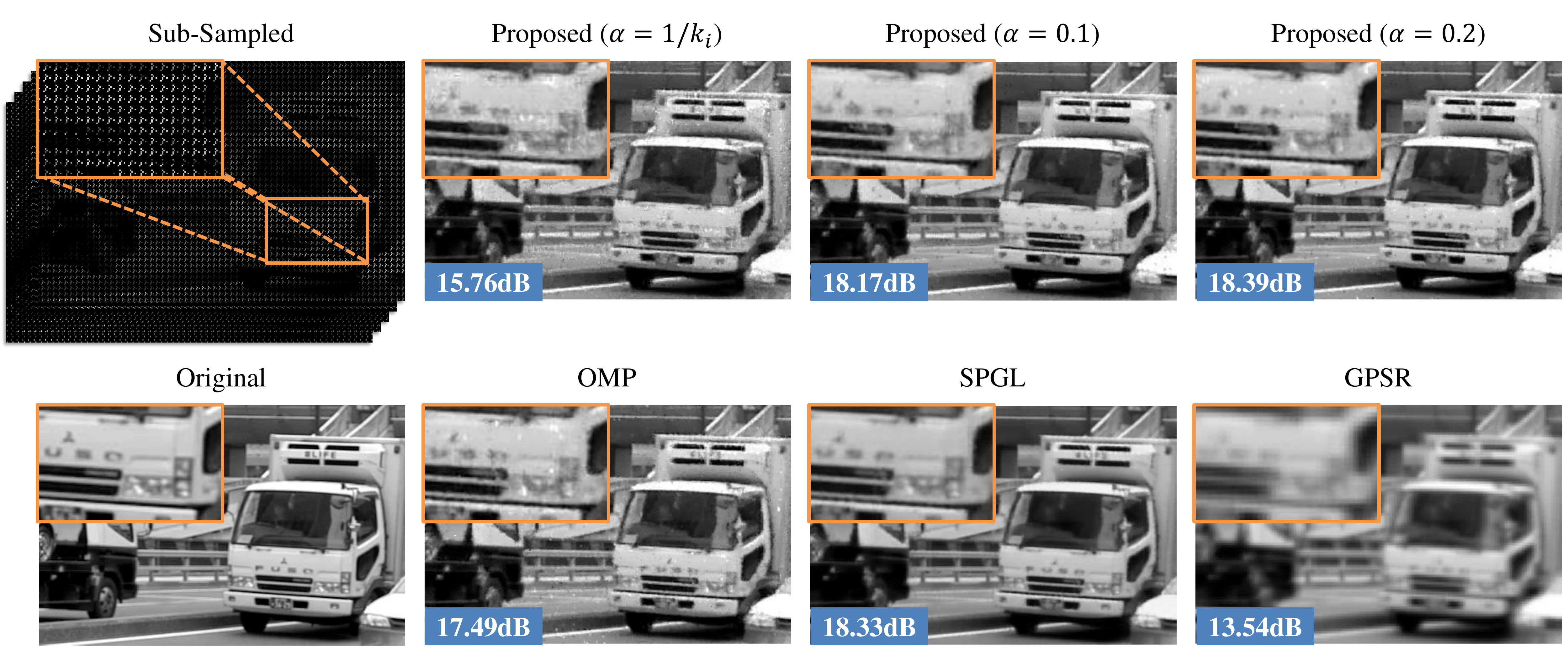}
\caption{Sample result for pixel-wise coded exposure compressive video sensing showing a frame from recovered video for our proposed method and OMP}\label{fig:VidSamp}
\label{fig:VidSamp}
\end{figure*}

\begin{table*}[!htb]
\begin{center}
\begin{tabular}{|l|c|c|c|c|c|c|c|c|c|c|c|}
\cline{3-12}
\multicolumn{1}{ l  }{}& \multicolumn{1}{ c|  }{}& \multicolumn{5}{|c|}{\bf{Denoising}}& \multicolumn{5}{|c|}{\bf{Video Compressive Sampling}}  \\
\cline{3-12}
\multicolumn{1}{ l  }{}& \multicolumn{1}{ c|  }{}& \multicolumn{2}{|c|}{\bf{Dogrun}}  & \multicolumn{2}{|c|}{\bf{Truck}} & \bf{Relative}& \multicolumn{2}{|c|}{\bf{Dogrun}}  & \multicolumn{2}{|c|}{\bf{Truck}} & \bf{Relative}\\
           
\cline{1-6}\cline{8-11}
&Dic.	& SNR      & Time                 & SNR      & Time      &    \bf{Run}   
				& SNR      & Time                 & SNR      & Time      &    \bf{Run}            \\

\bf{Method} &Size	& (dB)      & (min)                 & (dB)      & (min)       &    \bf{Time}   
				& (dB)      & (min)                 & (dB)      & (min)       &    \bf{Time}            \\

\hline\hline
Proposed ($\alpha = 1/k_{i}$) &100k   & 18.53         & 0.43                      & 18.76         & 0.17            & 1x
& 18.53         & 0.43                      & 18.76         & 0.17            & 1x             \\

Proposed ($\alpha = 0.1$)   &100k & 18.98         & 1.42                      & 19.10         & 0.49            &  3x      
& 18.24         & 4.00                      & 18.17         & 3.13            &  3x     \\

OMP &10k                     & 18.51         & 60.15                     & 18.36         & 23.84              &  140x     
& 16.97         & 38.33                    & 17.49         & 8.26              &  17x     \\


\hline
SPGL1  &10k              & 13.46         & 85.74                     & 19.21         & 105.87              & 411x   & 17.34         & 80.76                     & 18.33        & 45.49              & 50x       \\

GPSR  &10k               & 17.56         & 33.16                     & 15.80         & 17.03               & 88x    & 15.75         & 7.95                     & 13.54         & 23.05               & 14x      \\

\hline
KD-tree  &10k                 & 17.17             & 277.44                   & 16.15        & 137.78                    & 727x       & 16.48         & 17.60                   & 15.90       & 7.79                    & 10x       \\

ANN  &10k                    & 17.13         & 84.27                     & 16.10         & 36.28                      & 503x   & 16.46         & 13.76                     & 15.89         & 5.48                      & 7x   \\
\hline
\end{tabular}
\end{center}
\vspace{-0in}
\caption{Video Denoising and Pixel-wise coded exposure video sensing results: Our method has a run-time thats 100-500x faster with the same reconstruction quality.}
\label{table:VidDe}
\end{table*}

\textbf{Dictionary Construction:}
A general image dictionary is constructed from a set of 8 natural test images from the USC-SIPI Image Database (Barbara,  Boat, Couple, Elaine, House, Lena, Man, and Peppers) using $16\times 16$ patches and a shift of 2 pixels.   From each image, a dictionary of 5,000 atoms is learned using the K-SVD method.  These dictionaries were merged to create a 40,000 atom dictionary. This dictionary was  randomly sub-sampled to create a 4,000 element dictionary for use with FPC\_AS and SPGL1.
The dictionary was clustered using the hierarchical scheme of Section \ref{sec:Cluster} with 100 equally sized clusters in the first level ($k_1=100$), 10 equally sized cluster in the second level, and 10 equally sized clusters in the third (final) level ($k_2=k_3=10$).   Sparse coding is performed using Algorithm \ref{alg:match}.

\textbf{Image Denoising:}
Three images were selected from the Berkeley Segmentation Dataset image numbers 223061,102061, and 253027).  Each image was contaminated with Gaussian white noise to achieve an SNR of 10dB.
Greedy recovery was performed using 10 dictionary atoms per patch.
Sample denoising results are shown in Figure \ref{fig:ImgDe}.
Time trial results are shown in Table \ref{table:ImgDe}.

\textbf{Image Super-resolution:}
This experiment enhances the resolution of an image using information from a higher resolution dictionary.
We use three test images from the Berkeley Segmentation Database.
Low resolution images are broken into $4\times 4$ patches.
The low resolution  $4\times 4$ patches are mapped onto the $16\times 16$ dictionary patches for comparison, and then matched using sparse regularization algorithms \ref{alg:match} with sparsity $K=3$.
The reconstructed high-resolution patches are then averaged together to create a super-resolution image.
Sample super-resolution reconstructions are shown in Figure \ref{fig:ImgSupRes}.  Time trails are displayed in Table \ref{table:ImgDe}.

\Subsection{Video}

\textbf{Dictionary Construction:}
We obtained the dictionaries and high speed videos used in \cite{liu2013efficient} from the authors for this experiment.
MP experiments were done using a dictionary of $10^5$ atoms, and variational experiments were done using a randomly sub-sampled dictionary of $10^4$ atoms.   Video patches of size $7\times 7\times9$ are extracted from video frames.
The dictionary was clustered using the same parameters as the image dictionary.  Sparse coding is performed using Algorithm \ref{alg:match} with $\alpha = 0.1$.

\textbf{Video Denoising:}
Video denoising proceeds similarly to image denoising. The original 18 frame videos were contaminated with Gaussian white noise to have an SNR of 10dB. Patches of size $7\times7\times 18$ were extracted from the video to create test vectors of dimension $882$. For the ``dog'' video patches were generated with a shift of 1 pixel  (35673 patches) while for truck a 3 pixels shift was used (14505 patches). Sparse coding and recovery was performed using 10 atoms per patch.  Sample frames from denoised videos are shown in Figure \ref{fig:VidDe} and runtimes are displayed in Table \ref{table:VidDe}.

\textbf{Video Compressive Sampling:}
We emulate the video compressive sampling experiments in \cite{liu2013efficient}.
This experiment simulates a pixel-wise coded exposure video camera much like \cite{liu2013efficient}\cite{reddy2011p2c2}.
The pixel-wise coded exposure video camera operates at $\frac{1}{9}$ the frame-rate of the reconstructed video and therefore results in $\frac{1}{9}$ samples/measurements compared to the original video.
For reconstruction, we closely follow the approach of \cite{liu2013efficient} and reconstruct the video by using patch-wise sparse coding using the learned dictionary. Sample frames from reconstructed videos are shown in Figure \ref{fig:VidSamp} and runtimes are displayed in Table  \ref{table:VidDe}.

\Subsection{Light Field Analysis}

\textbf{Dictionary Construction:}
A dictionary was created for light field patches using several sample light fields:  synthetic light fields created from the ``Barbara'' test image as well as several urban scenes and light field data from the MIT Media Lab Synthetic Light Field Archive (Happy Buddha, Messerschmitt, Dice, Green Dragon, Mini Cooper, Butterfly, and Lucy).  Dictionaries are learned on 4-dimensional patches that consist of an 8x8 pixel grid and a 5x5 view window (total dimensions per patch is 8x8x5x5 = 1600). By combining patches from all training data, a dictionary with $146,000$ atoms was built.  Again we randomly sampled the dictionary to generate a small $10,000$ atom dictionary for methods that were intractably slow when using the full-sized dictionary.

The dictionary was  subjected to hierarchical clustering using the same parameters as the image dictionary, with 100 equally sized clusters in the first level, 10 equally sized cluster in second level, and 10 equally sized clusters in third (final) level.   Sparse coding is performed using Algorithm \ref{alg:match} with $\alpha = 0.1$.

\begin{figure}[!t]
   \centering
      \includegraphics[scale = 0.2]{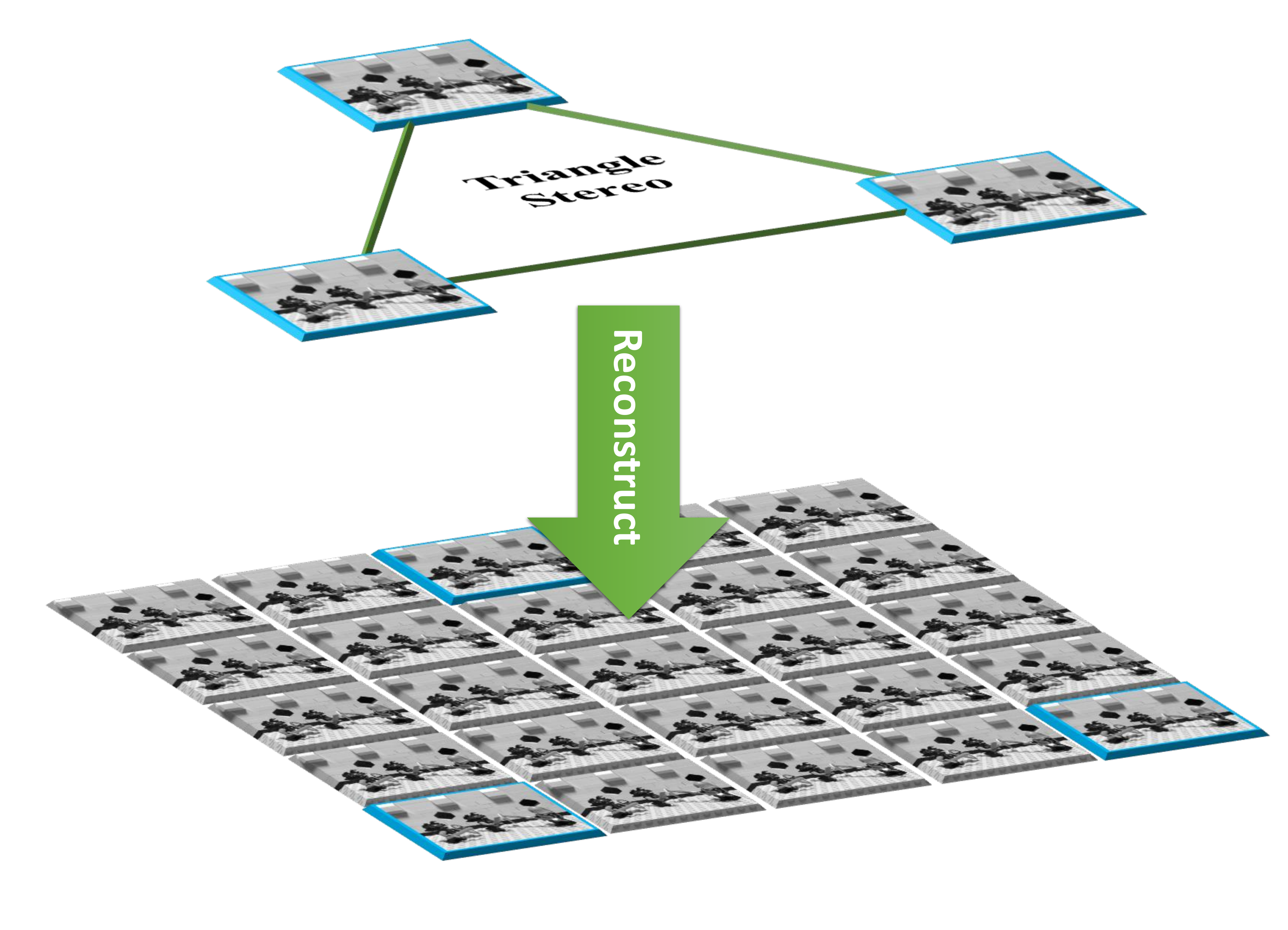}
   \caption{Light Field from Trinocular Stereo: $5\times5$ view light field is recovered using only three views. }\label{fig:TrStereo}
\label{fig:LFTr}
\end{figure}

\textbf{Light-Field Denoising:}
Denoising experiments were performed using the ``Tarot Cards'' and ``Crystal Ball'' datasets from the Stanford Light Field Archive. We add noise to the light field to achieve an SNR of 10dB. Patches are extracted with a 2 pixels shift (15625 patches).  Because of the high dimensionality of light-field patches, sparse coding was done using a sparsity of 50.
Results are displayed in Figure \ref{fig:LFDe} and Table  \ref{table:LFDe}.

\begin{figure*}[!htb]
\centering
\includegraphics[scale = 0.4]{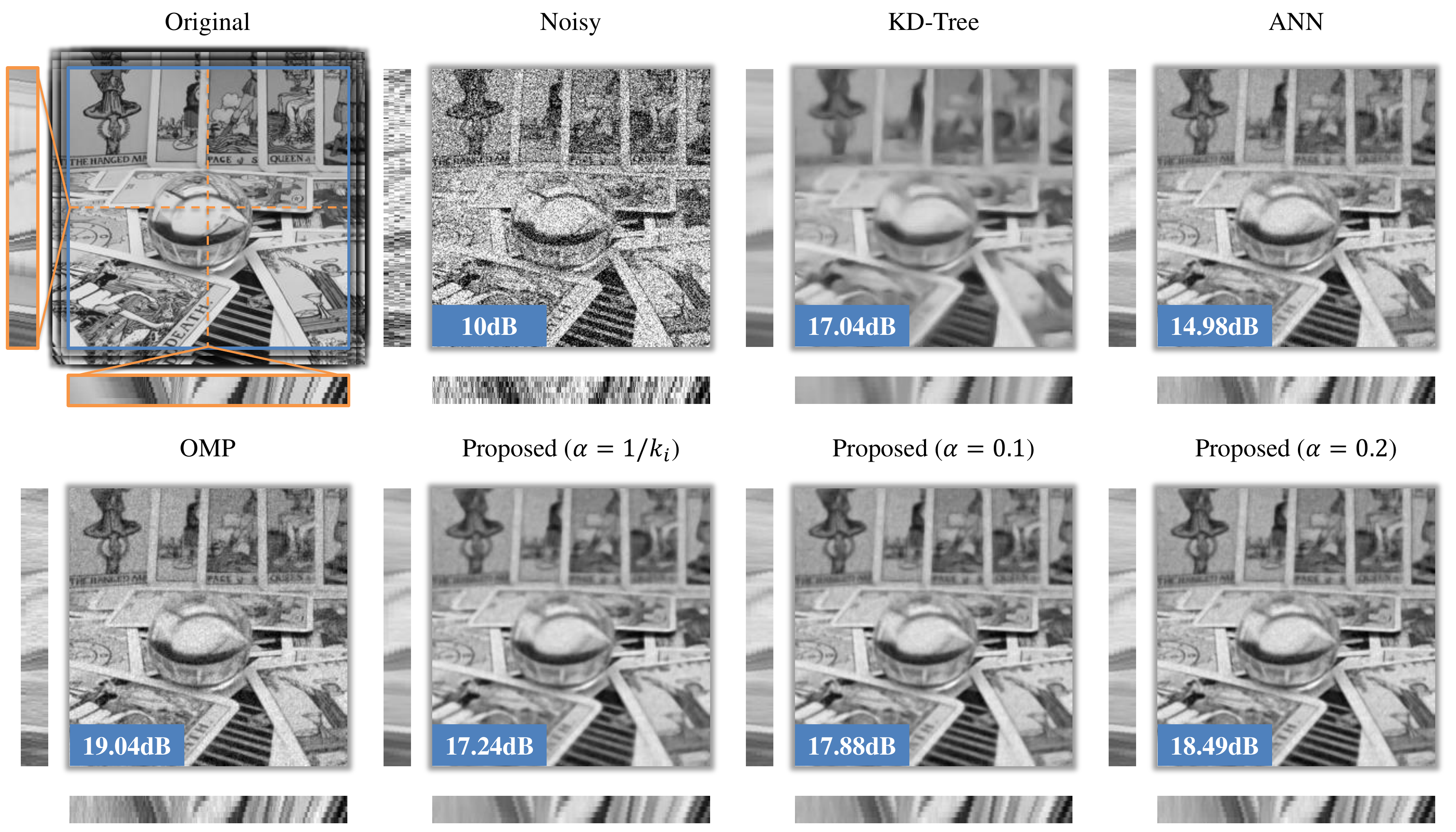}
\caption{Sample result for Light Field Denoising showing center view.}\label{fig:LFDe}
\label{fig:LFDe}
\includegraphics[scale = 0.4]{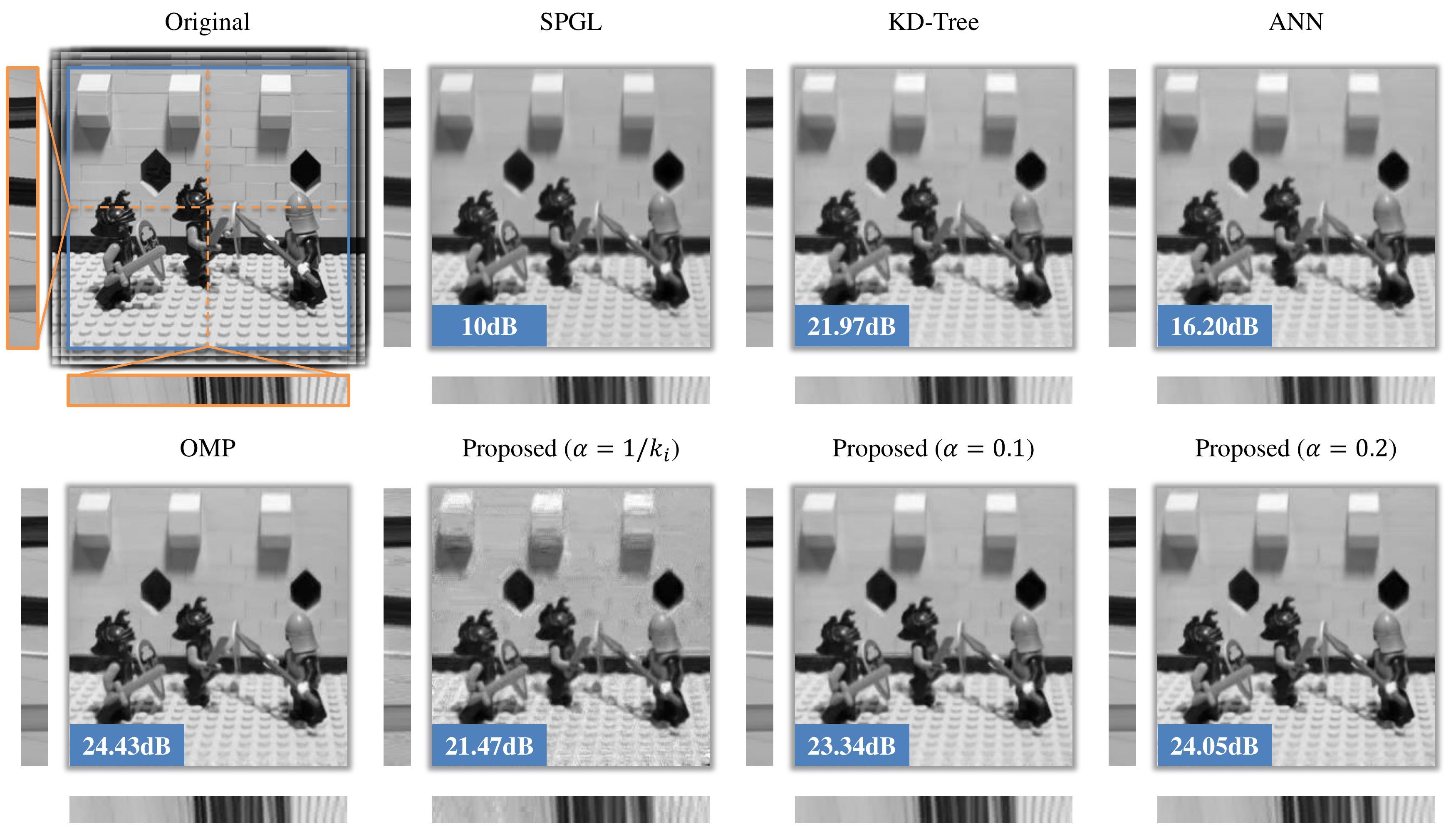}
\caption{Sample of light field reconstructed from 3 views. Shown is the middle view. }\label{fig:LFTr}
\label{fig:LFTr}
\end{figure*}

\begin{table*}[!htb]
\begin{center}
\begin{tabular}{|l|c|c|c|c|c|c|c|}
\cline{3-8}
\multicolumn{1}{ l  }{}& \multicolumn{1}{ c|  }{}& \multicolumn{3}{|c|}{\bf{Denoising}}& \multicolumn{3}{|c|}{\bf{Light-Field from Trinocular Stereo}}  \\
\hline
            &Dic.            & \multicolumn{2}{|c|}{\bf{Tarot}}  &    \bf{Relative}  & \multicolumn{2}{|c|}{\bf{Tarot}}  &    \bf{Relative}\\
\cline{3-4}\cline{6-7}
\bf{Method} &Size            & SNR(dB)      & Time(min)               &    \bf{Run Time}          & SNR(dB)      & Time(min)               &    \bf{Run Time} \\
\hline\hline
Proposed ($\alpha = 1/k_{i}$) &146k   & 17.24         & 2.10                    & 1x      & 21.47         & 0.09                    & 1x       \\
Proposed ($\alpha = 0.1$) &146k   & 17.88         & 4.92                       &  2x      & 23.34         & 0.25                       &  2x     \\
\hline
OMP &10k                     & 19.04         & 214.36                      &  102x    & 24.43         & 86.89                      &  965x      \\
\hline
SPGL1  &10k              & 10.71         & 83.86                        & 40x      & 22.28         & 104.06                        & 1156x     \\
GPSR  &10k               & 13.24         & 24.47                         & 12x      & 20.48         & 27.26                         & 302x     \\
\hline
KD-Tree &10k                & 17.04         & 401.56                 & 191x      & 21.97         & 32.73                  & 363x         \\
ANN  &10k                   & 14.98         & 6.63          & 3.15x   & 16.20         & 13.88          & 154x    \\
\hline
\end{tabular}
\end{center}
\vspace{-0in}
\caption{Reconstruction of noisy Light Field for $5\times5$ views and comparison to other methods and light field reconstruction from 3 views. The proposed method obtain similar quality in much shorter run-time. }\label{table:LFDe}
\end{table*}

\textbf{Light-Field from Trinocular Stereo:}
In this experiment, we attempt to reconstruct a light field with $5\times5$ views from just three cameras (trinocular), much like \cite{mitra2012light}.
The Lego Knights light field dataset from the Stanford Light Field Archive we subsampled to retain only the top middle, bottom left, and bottom right views of the $5\times5$ view grid at each pixel.   Patches of size $8\times8\times5\times5$ were then sampled with 2 pixel shift.  The observed patch data was mapped onto the corresponding entries for each dictionary atom, and used for sparse coding. This reduces the dimension of the test set and dictionary from 1600 to 1600x(3/25)= 192. Sparse coding was performed with 10 dictionary atoms per patch. Restored patches were then averaged to reconstruct the full light field with $5\times 5$ views.  Results are displayed in Figure \ref{fig:LFTr} and  Table \ref{table:LFDe}.

\Section{Discussion and Conclusions}
\label{sec:Discussions}

  The high performance of shallow trees for dictionary matching seems to contradict the conventional intuition that deeper tree are better.  For image dictionaries, it seems that atoms are naturally organized into a large number of separated clusters with fairly uniform separation.  By exploiting this structure at a high level, shallow trees perform highly accurate matching using relatively few comparisons.  In contrast, deep tree nearest neighbor searches require a smaller number of dot products to descend to the bottom of the tree.  However, these approaches require branch-and-bound methods that backtrack up the tree and explore multiple branches in order to achieve an acceptable level of accuracy.   For well clustered data such as the dictionaries considered here, the shallow tree approach achieves superior performance by avoiding the high cost of backtracking searches through the tree.

{\small
\bibliographystyle{IEEEtran}
\bibliography{IEEEabrv,egbib}
}

\end{document}